\documentclass{article}

 \usepackage[preprint]{neurips_2026}

\usepackage[utf8]{inputenc}
\usepackage[T1]{fontenc}
\usepackage[hidelinks]{hyperref}
\usepackage{url}
\usepackage{graphicx}
\usepackage{amsmath}
\usepackage{amssymb}
\usepackage{multirow}
\usepackage{svg}
\usepackage{tabularx}
\usepackage{booktabs}
\usepackage{array}
\usepackage[capitalize,noabbrev]{cleveref}
\usepackage[table]{xcolor}
\usepackage{colortbl}
\usepackage{subcaption}
\usepackage{microtype}
\usepackage{enumitem}
\usepackage{placeins}
\setlist{nosep}
\usepackage{orcidlink}

\definecolor{lightgray}{gray}{0.9}

\newcommand{\parag}[1]{\vspace{-3mm}\paragraph{#1}}

\usepackage[noblocks]{authblk}

\title{Learning to Generate Multiple Objects from Dense and Occluded Layouts}

\author[1,2$\dagger$]{Bach-Hoang Ngo\orcidlink{0009-0002-2290-1187}}

\author[1,2$\dagger$]{Si-Tri Ngo\orcidlink{0009-0008-6439-4850}}

\author[3$\ddagger$]{Hieu Le\orcidlink{0000-0001-7855-2778}}

\author[1,2$\ddagger$*]{Trung-Nghia Le\orcidlink{0000-0002-7363-2610}}

\affil[1]{University of Science, Ho Chi Minh City, Vietnam}

\affil[2]{Vietnam National University, Ho Chi Minh City, Vietnam}

\affil[3]{UNC Charlotte, North Carolina, United States}

\begin{document}
\maketitle
\vspace{-8mm}
\begin{center}
\href{https://bachngoh.github.io/AIBL/}{\texttt{https://bachngoh.github.io/AIBL/}}
\end{center}
\vspace{1mm}

\renewcommand{\thefootnote}{\fnsymbol{footnote}}
\footnotetext[0]{$^\dagger$Equal contributions \quad $^\ddagger$Equal advising \quad $^*$Corresponding author} 
\footnotetext[0]{Emails: \texttt{nhbach22@apcs.fitus.edu.vn}, \texttt{ntsi22@apcs.fitus.edu.vn}, \texttt{hle40@charlotte.edu}, \texttt{ltnghia@fit.hcmus.edu.vn}} 
\renewcommand{\thefootnote}{\arabic{footnote}}

\begin{abstract}
Text-to-image diffusion models fail to generate correct object counts in dense scenes, where overlapping instances collapse into indistinguishable structures despite appearing visually plausible. We identify this as instance ownership collapse: tokens from overlapping objects interact freely through attention, while heavily occluded instances receive weak supervision due to their small visible areas.
We address this through layout-aware attention biases that softly bias token interactions toward region-consistent grouping and suppress cross-instance leakage, paired with an amodal-balanced loss that amplifies gradients for occluded objects based on their occlusion level. To enable systematic evaluation, we introduce OverlapDepth-45K, a benchmark of densely overlapping scenes with amodal supervision. Our approach substantially improves count accuracy and prevents instance merging while preserving image quality.
\end{abstract}

\section{Introduction}
\label{sec:intro}

\begin{figure*}[t]
  \centering
  \includegraphics[width=0.8\textwidth]{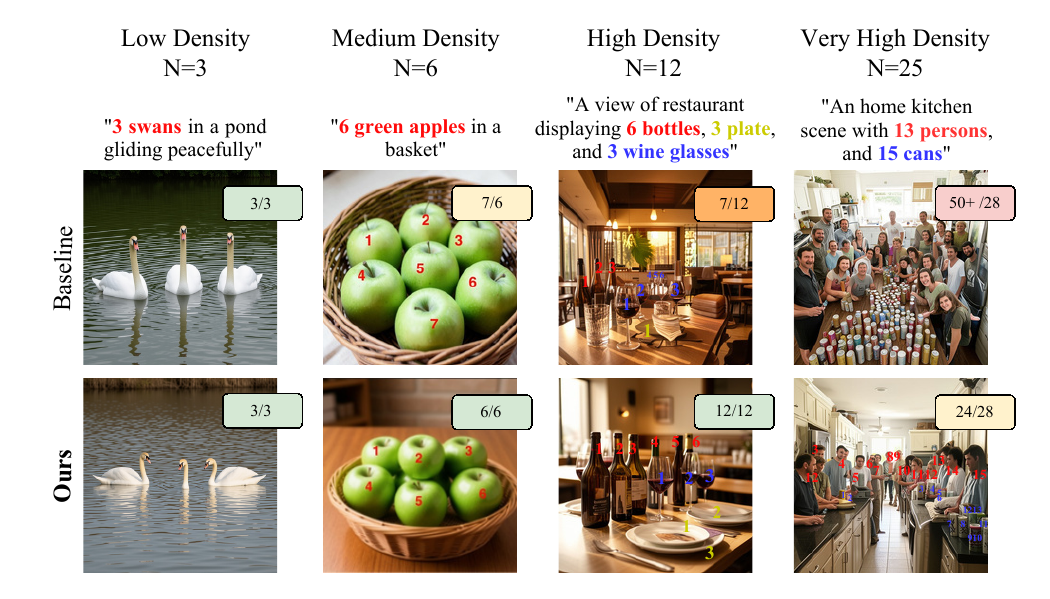}
  \vspace{-3mm}
  \caption{\textbf{Count-preserving multi-category compositional generation.} Standard diffusion models (top) succeed on simple prompts (3/3 swans) but fail as scene complexity grows - overcounting single-category scenes (7/6 apples), dropping instances in multi-category compositions (7/12 across bottles, plates, and glasses), and catastrophically merging objects in dense scenes (50+/28 across persons and cans). Our approach (bottom) achieves exact or near-exact counts across all density levels (3/3, 6/6, 12/12, 24/28) while preserving photorealistic quality and natural inter-object occlusion.}
  \vspace{-5mm}
  \label{fig:teaser}
\end{figure*}

Generating correct content is a fundamental requirement for controllable image synthesis. Yet modern text-to-image (T2I) models remain unreliable when asked to produce scenes with explicit object counts~\cite{blackforestlabs2024announcingbfl, dalle3, podell2024sdxl}. Prompts specifying dozens of instances frequently result in missing objects, merged entities, or visually plausible but numerically incorrect compositions.
As shown in \cref{fig:teaser}, while a standard model handles simple prompts (3 swans), it overcounts single-category scenes (7 instead of 6 apples), drops instances in multi-category compositions (7 instead of 12 objects), and catastrophically merges dense scenes (50+ instead of 28 across persons and cans).
This limitation significantly hinders the utility of T2I models in applications demanding high structural fidelity.

Free-form generation without structural priors rarely satisfies cardinality constraints. Conditional generation with layout or mask priors partially alleviates this problem by anchoring objects spatially~\cite{gligen, zheng2023layoutdiffusion}. However, once scenes become dense and objects heavily overlap, failures persist. Even with correct bounding boxes, instances collapse into indistinguishable blobs: the model may allocate roughly the correct semantic mass within a region, but individual objects are not rendered as separable, countable entities.

We argue that this failure arises from how diffusion transformers represent and learn overlapping objects. In dense layouts, tokens from different instances frequently occupy nearby or overlapping spatial regions. Standard self-attention allows features to freely propagate across these regions, causing representations of adjacent objects to blend together during generation. Meanwhile, the diffusion training objective applies a uniform reconstruction loss over all pixels, which implicitly favors large and fully visible regions. When objects are heavily occluded, only a small fraction of their pixels contributes to the loss, producing weak gradients for those instances. Over training, this imbalance teaches the model to allocate most of its capacity to visible regions while neglecting partially occluded objects. Consequently, when layouts overlap, the model tends to synthesize shared textures across neighboring boxes rather than maintaining distinct instance boundaries.

To address this, we introduce a framework that encourages instance ownership throughout generation and training. We inject layout-aware attention biases into the transformer to softly favor region-consistent token grouping, while simultaneously applying an amodal-balanced loss that amplifies gradients for occluded instances based on the ratio between their full and visible extents. These two mechanisms operate together: layout attention improves instance separability in dense overlaps, and the balanced objective ensures that occluded objects remain strong optimization targets during training.

Concretely, our \textit{layout-aware attention} modifies the attention logits of the diffusion transformer using region identities derived from the input layout. Tokens belonging to the same bounding box receive a positive cohesion bias that encourages intra-instance interaction, while tokens from different boxes receive a separation bias that suppresses cross-instance feature leakage. This simple modification provides soft spatial guidance to reduce local blending when regions overlap, rather than enforcing hard geometric adherence to box boundaries. Complementing this structural constraint, the \textit{amodal-balanced loss} adjusts the diffusion training objective to account for occlusion. During training, we estimate each instance's occlusion level from its amodal and visible masks and compute a scaling factor proportional to the ratio between its full extent and visible area. This factor is applied to the reconstruction loss of pixels owned by the instance, ensuring that partially visible objects contribute gradients comparable to fully visible ones.

To systematically study dense compositional generation under severe overlap, we further introduce \textbf{OverlapDepth-45K}, a dataset of highly populated scenes with scalable amodal supervision derived from standard instance annotations. This dataset provides controlled evaluation of object counting and instance separability across a wide range of occlusion levels.

In summary, our key contributions are:
\begin{enumerate}
    \item \textbf{Ownership-Aware Layout Attention:} We introduce a parameter-free attention bias for DiT architectures that exploits the depth-ordered ownership structure of overlapping bounding boxes. By reinforcing cohesion for ownership-winner tokens in contested zones and strengthening separation between overlapping instances, the mechanism prevents cross-instance feature leakage without rigid hard-masking or additional trainable parameters.
    \item \textbf{Amodal-Aware Instance-Balanced Loss (AIBL):} We propose a novel training-time loss reweighting scheme that amplifies gradients for occluded instances using amodal-to-visible area ratios, improving counting fidelity in dense scenes without introducing inference-time mask dependencies.
    \item \textbf{Empirical Validation:} We demonstrate that layout attention combined with AIBL effectively mitigates catastrophic object fusion, reducing RMSE by 35.3\% (from 9.26 to 5.99) over the layout-only baseline on our OverlapDepth-45K benchmark while maintaining high precision on tolerance thresholds. Our proposed approach significantly outperforms existing methods on T2I-Compbench~\cite{t2icompbench} and CoCoCount~\cite{makeitcount} datasets.
\end{enumerate}

\section{Related Work}
\parag{Counting in T2I.}
Modern text-to-image diffusion models still struggle with precise counting, especially as object counts increase~\cite{t2icompbench}. Existing solutions include test-time guidance from auxiliary counting networks~\cite{counting_guidance}, resampling-based correction methods such as CountGen~\cite{makeitcount}, and training losses that encourage stronger token-object alignment~\cite{chefer2023attend}. These methods improve sparse scenes but often scale poorly to dense layouts. A key limitation is that most counting or attention losses aggregate activations globally or inside coarse boxes, without distinguishing visible from occluded object regions. As a result, overlapping objects can be incorrectly penalized as missing, causing unstable gradients in dense scenes. We instead introduce an amodal-aware objective that reweights instance contributions by visibility, improving high-density counting without expensive test-time optimization.

\parag{Layout-to-image generation and spatial control.}
Layout-to-image methods reduce prompt ambiguity by conditioning generation on spatial priors. GLIGEN~\cite{gligen}, ControlNet~\cite{zhang2023controlnet}, T2I-Adapter~\cite{mou2023t2i}, LayoutDiffusion~\cite{zheng2023layoutdiffusion}, and later methods such as InstanceDiffusion~\cite{wang2024instancediffusion}, MIGC~\cite{zhou2024migc}, and 3DIS-FLUX~\cite{3dis2025} improve object placement through grounded tokens, attention constraints, decoupled decoding, or depth-aware control. Training-free layout guidance has also been explored through cross-attention manipulation~\cite{chen2024trainingfree, phung2023grounded}. However, layout conditioning mainly provides a soft placement prior. Under severe box overlap, standard layout guidance can still collapse multiple instances into one blended object. In our framework, layout attention anchors object locations, while the proposed amodal-aware loss directly addresses the dense instance-fusion problem.

\parag{Occlusions and dense overlaps.}
Severe occlusion remains difficult for compositional generation. SeeThrough3D~\cite{agrawal2026seethrough3docclusionaware3d} models occlusion-aware 3D boxes, while OMG~\cite{kong2024omg} uses a two-stage layout and comprehension pipeline to reduce missing entities. Recent benchmarks such as OverLayBench~\cite{li2025overlaybench} further show that amodal masks alone are insufficient for dense overlaps. The core issue is not only knowing what object shapes exist, but deciding how training gradients should be assigned when instances overlap. Our Amodal-Aware Instance-Balanced Loss (AIBL) addresses this optimization imbalance by weighting instances according to occlusion degree, leading to stronger count preservation in highly dense scenes.

\begin{figure*}[t]
  \centering
    \includegraphics[width=\textwidth]{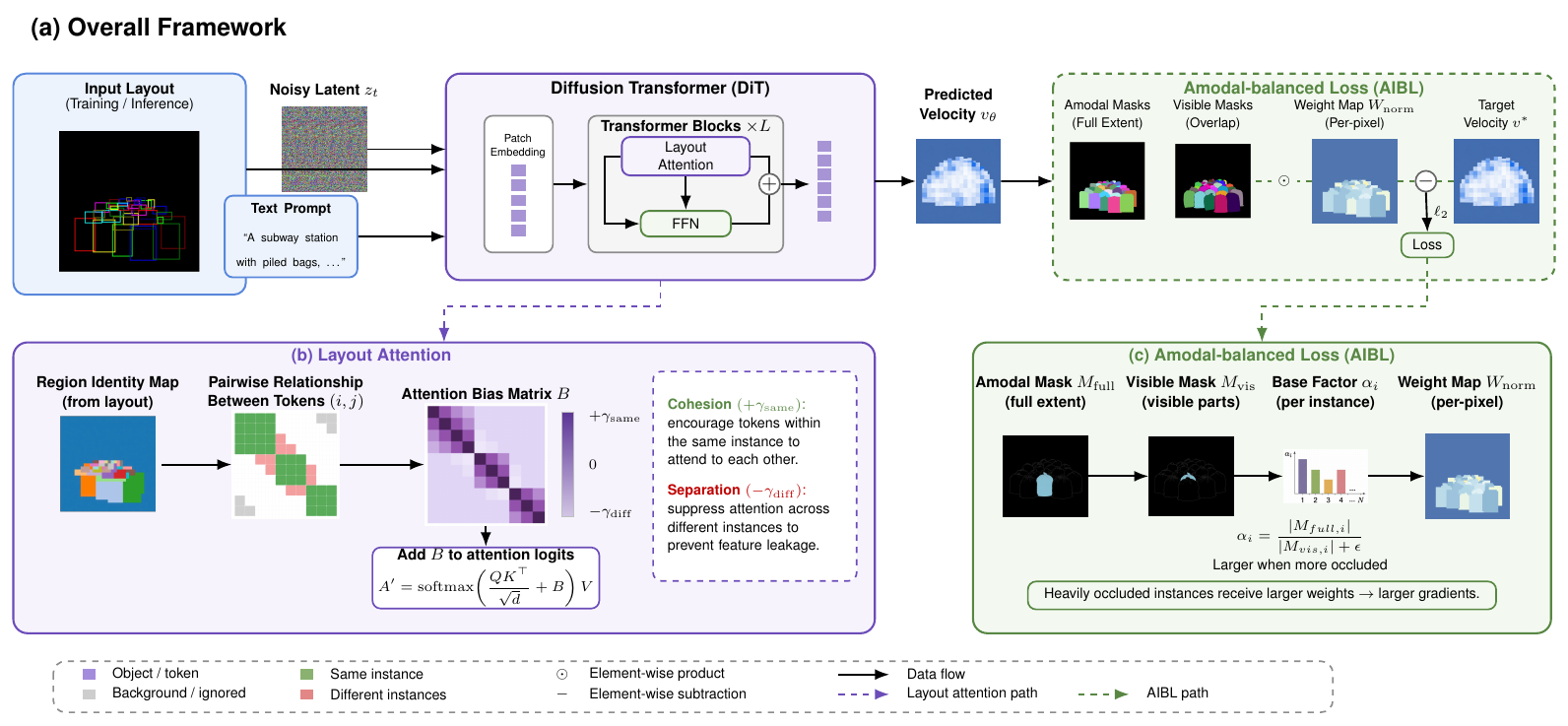}
  \vspace{-3mm}
  \caption{\textbf{Architecture Overview.} (a) Layout Attention (LA) is integrated into the transformer blocks of a DiT backbone to steer the generation process. (b) LA utilizes a region identity map to compute an attention bias matrix $M$, where cohesion and separation biases are applied to tokens based on their instance ownership. (c) The Amodal-Aware Instance-Balanced Loss (AIBL) module (training only) computes per-instance occlusion boost factors $\alpha_i$ from amodal and visible masks. These factors produce a mean-normalized weight map $W_{norm}$ that prioritizes heavily occluded objects during training.}
  \vspace{-5mm}
  \label{fig:pipeline}
\end{figure*}

\section{Proposed Method}
\label{sec:method}
Generating correct object counts in dense scenes fails not because
diffusion models lack spatial awareness, but because their training
objectives are blind to occlusion. Under standard uniform reconstruction
losses, heavily occluded instances contribute negligible gradients,
effectively teaching the model to drop them. Layout conditioning alone
cannot fix this: even with correct bounding boxes, the optimization
imbalance persists.

We address this with two complementary components (see ~\cref{fig:pipeline}): (1) a
\textbf{Layout Attention mechanism} that injects region-aware spatial
biases into the transformer's attention logits, softly encouraging
region-consistent interactions at both training and inference without additional
trainable parameters; and (2) the \textbf{Amodal-Aware Instance-Balanced
Loss (AIBL)}, which scales each instance's loss contribution in inverse
proportion to its visibility, restoring balanced gradient flow without
introducing any inference-time mask dependency.


\subsection{Layout Attention}
\label{ssec:layout_attn}

Rather than introducing separate trainable cross-attention modules or manipulating latents at inference via gradient guidance, we inject soft region priors directly within the existing attention mechanism of the diffusion transformer. Our layout attention requires no additional learnable parameters beyond the LoRA adapters and operates identically at training and inference, permanently closing any training-inference gap.

\textbf{Region identity map.} Given a text prompt $\mathcal{P}$ with $N$ target instances, we define a layout as $\mathcal{B} = \{b_1, b_2, \ldots, b_N\}$, where each $b_i = (x_i, y_i, w_i, h_i) \in [0, 1]^4$ specifies the normalized spatial coordinates for the $i$-th instance. We construct a discrete \textit{region identity map} $\mathcal{R} \in \mathbb{Z}^{H_l \times W_l}$ over the latent spatial grid by painting bounding boxes in depth order (i.e., a painter's algorithm): each token is assigned the region ID of the \textit{frontmost} instance whose box covers it, and background tokens receive an ID of 0. This depth-ordered assignment is the key structural prior that our attention bias exploits.

\textbf{Overlap-ownership attention bias.} In dense layouts, bounding boxes inevitably overlap. When two boxes intersect, the painter's algorithm assigns contested tokens exclusively to the depth-front instance---the \textit{ownership winner}. The occluded instance retains only its non-overlapping tokens, which may be a small fraction of its total extent. This asymmetry creates a natural \textit{ownership structure} over the latent grid: tokens in overlapping regions belong unambiguously to a single instance, but they are spatially adjacent to tokens of the occluded object and thus especially prone to cross-instance feature leakage through attention.

We exploit this structure by computing a per-token \textit{coverage count} $c(p) = |\{b_i \in \mathcal{B} \mid p \in b_i\}|$ and defining the overlap zone $\mathcal{Z} = \{p \mid c(p) \geq 2\}$. The attention bias matrix $M \in \mathbb{R}^{S \times S}$ (where $S = S_{\text{txt}} + H_l W_l$) over all pairs of image tokens is then:

\begin{equation}
  M(p_i, p_j) =
  \begin{cases}
    \beta \cdot \omega_{\text{intra}} + \Delta_{\text{coh}}(p_i, p_j) & \text{if } \mathcal{R}(p_i) = \mathcal{R}(p_j) > 0 \\[3pt]
    \omega_{\text{inter}} + \Delta_{\text{sep}}(p_i, p_j) & \text{if } \mathcal{R}(p_i) \neq \mathcal{R}(p_j),\  \text{both} > 0 \\[3pt]
    0 & \text{otherwise}
  \end{cases}
  \label{eq:attn_bias}
\end{equation}
where $\omega_{\text{intra}} > 0$ is the base intra-instance cohesion bias, $\omega_{\text{inter}} < 0$ is the base cross-instance separation bias, and $\beta > 1$ is a global grounding factor for dense scenes. The overlap-adaptive corrections $\Delta_{\text{coh}}$ and $\Delta_{\text{sep}}$ activate only in the presence of contested tokens. Let $\mathcal{Z}_k = \{p \mid \mathcal{R}(p) = k,\, p \in \mathcal{Z}\}$ denote the set of overlap-zone tokens owned by instance $k$. The corrections are:


\begin{equation}
\small
\begin{aligned}
\Delta_{\text{coh}}(p_i, p_j) &=
\begin{cases}
\tfrac{1}{2}\beta\omega_{\text{intra}} 
& p_i, p_j \in \mathcal{Z}_k \text{ for some } k,\\
0 & \text{otherwise,}
\end{cases}
\end{aligned}
\quad
\begin{aligned}
\Delta_{\text{sep}}(p_i, p_j) &=
\begin{cases}
\tfrac{1}{2}\omega_{\text{inter}} 
& \mathcal{Z}_{\mathcal{R}(p_i)} \neq \emptyset 
\text{ or } \mathcal{Z}_{\mathcal{R}(p_j)} \neq \emptyset,\\
0 & \text{otherwise.}
\end{cases}
\end{aligned}
\label{eq:overlap_corrections}
\end{equation}

The cohesion correction $\Delta_{\text{coh}}$ reinforces the identity of ownership-winner tokens that lie in the contested zone, preventing their features from being ``pulled'' toward the occluded instance. The separation correction $\Delta_{\text{sep}}$ operates at the \textit{region} level: once any instance has tokens in a contested area, \textit{all} cross-instance pairs involving that instance receive stronger repulsion. This whole-region propagation is deliberate---it prevents leakage not just at the overlap boundary but along the entire interface between neighboring instances, which is where instance merging initiates. Both corrections scale the existing bias magnitudes by a fixed factor of $\frac{1}{2}$, which we do not tune across experiments.

\textbf{Gaussian soft prior.} To complement the discrete region bias with smooth spatial guidance, we augment $M$ with a distance-weighted Gaussian centered at each bounding box's centroid:
\begin{equation}
  M_{\text{soft}}(p_j, b_i) = \exp\!\left(-\frac{\|\text{pos}(p_j) - \text{center}(b_i)\|^2}{2\sigma^2}\right).
  \label{eq:soft_attn}
\end{equation}
This provides a gentle spatial decay that softens the hard region boundaries, encouraging tokens near a box's center to attend more strongly to their assigned instance while gradually releasing peripheral tokens.

The combined bias $M + M_{\text{soft}}$ is injected as an additive term into the pre-softmax attention logits of the DiT's joint attention blocks. Because this modifies only the attention distribution rather than introducing new trainable pathways, layout attention preserves the pretrained generative prior while providing ownership-aware soft region guidance for count-faithful instance separation.

\subsection{Amodal-Aware Instance-Balanced Loss (AIBL)}
\label{ssec:aibl}

Standard diffusion models minimize a uniform Mean Squared Error (MSE) over all spatial
positions equally. This creates a severe optimization imbalance in occluded objects. Consider a
training image containing two objects: a large foreground sofa occupying 40\%
of the image and a heavily occluded cushion with only 5\% of its area visible.
The sofa generates roughly $8 \times$ more gradient signal than the cushion
under uniform MSE, simply because it occupies more pixels. Over training, the model learns that reconstructing the occluded object has little impact on the loss. As a result, occluded instances are often merged or omitted at inference.

\textbf{Occlusion boost factor.} AIBL addresses this by leveraging amodal masks \textit{exclusively to reweigh the loss function}, ensuring the model receives amplified error signals for occluded instances without conditioning on masks at inference. For each instance $i$ in a training sample, we compute a multiplicative occlusion boost factor:

\begin{equation}
   \alpha_i = \min\left( \alpha_{max}, \left( \frac{A_i^{amodal}}{A_i^{visible} + \epsilon} \right)^\gamma \right),
   \label{eq:alpha}
\end{equation}
where $A_i^{amodal}$ is the instance's theoretical full (unoccluded) area, $A_i^{visible}$ is its actually visible area in the composite, $\gamma \in (0, 1)$ is a controlling parameter, and $\alpha_{max}$ clips extreme values. Intuitively, a fully visible object receives $\alpha_i \approx 1$, while a 90\%-occluded object receives $\alpha_i \gg 1$.

\paragraph{Pixel-wise weight map construction.}
Instance-level boost factors must be translated into a spatial weight map
that can be applied to the per-pixel reconstruction loss. We construct
$W \in \mathbb{R}^{H \times W}$ by assigning each pixel the boost factor
$\alpha_i$ of its owning instance, where ownership is determined by the
depth-ordered ownership map $\mathcal{O}$ provided by the dataset (i.e.,
the frontmost instance at each pixel is its owner):

\begin{equation}
    W(x, y) =
    \begin{cases}
        \alpha_{\mathcal{O}(x,y)} & \text{if } \mathcal{O}(x, y) > 0 \\
        \lambda_{bg}              & \text{if } \mathcal{O}(x, y) = 0
    \end{cases}
    \label{eq:weight_map}
\end{equation}

\noindent where $\mathcal{O}(x, y) \in \{0, 1, \ldots, N\}$ denotes the
instance ID of the frontmost object at pixel $(x, y)$, with $0$ reserved
for background. Background pixels receive a downweighted baseline value
$\lambda_{bg} = 0.8$, reflecting that background regions carry less
instance-specific information and should not dominate the loss even when
they are large.

\paragraph{Mean normalization.}
Applying per-instance boosts naively would inflate the overall loss
magnitude, implicitly increasing the effective learning rate in dense
scenes. To prevent this, we mean-normalize $W$ across all spatial
positions within each training sample:

\begin{equation}
    W_{norm}(x, y) = \frac{W(x, y)}{\frac{1}{HW}\sum_{x',y'} W(x', y')}.
    \label{eq:wnorm}
\end{equation}

\noindent This ensures that the sum of all pixel weights equals $HW$ fore
every training image, exactly preserving the global gradient scale
regardless of scene complexity. A heavily occluded scene and a simple
scene therefore produce gradients of the same expected magnitude, and the
learning rate requires no adjustment.

\subsection{Training Objective}
\label{ssec:objective}

The complete architecture is trained end-to-end using rectified flow matching~\cite{lipman2023flow}. Given a clean latent $\mathcal{Z}_0$, a noise sample $\epsilon$, and a random timestep $t \sim \mathcal{U}(0,1)$, the noisy latent is formed as $\mathcal{Z}_t = (1-t)\mathcal{Z}_0 + t\epsilon$ with target velocity $v_{target} = \epsilon - \mathcal{Z}_0$. The AIBL-weighted training objective is:

\begin{equation}
   \mathcal{L}_{total} = \mathbb{E}_{\mathcal{Z}_0, \epsilon, t, \mathcal{P}, \mathcal{B}} \left[ W_{norm} \odot \| v_{\theta}(\mathcal{Z}_t, \mathcal{P}, \mathcal{B}) - v_{target} \|_2^2 \right],
   \label{eq:loss}
\end{equation}
where $v_{\theta}$ denotes the model's velocity prediction conditioned on the layout-biased attention. To match the diffusion latent space, the pixel-space amodal and ownership masks are downsampled to the latent resolution via interpolation prior to computing the weight map $W_{norm}$. When AIBL is disabled, $W_{norm} = \mathbf{1}$ and the objective reduces to standard uniform MSE. Layout attention bias $M$ (Eq.~\ref{eq:attn_bias}) is applied identically during both training and inference, ensuring no training-inference gap.

\section{Experiments}
\label{sec:experiments}

\subsection{OverlapDepth-45K Dataset}

Training a depth-aware, layout-guided model requires a large-scale dataset featuring reliable amodal bounding boxes and pairwise ordinal depth metadata, particularly in highly occluded scenarios. Existing datasets like MS-COCO~\cite{lin2014coco} or Visual Genome~\cite{krishna2017vg} lack the dense, strictly controlled occlusion complexity required to elicit overlap ambiguity. To address this, we introduce the \textit{OverlapDepth-45K} dataset.

Our OverlapDepth-45K dataset is a high-overlap compositional benchmark containing 45,000 images. The dataset is split into 44,000 training images and 1,000 test images, with scenes densely populated by an average of 13.2 objects per image (typically 6-20 objects). It spans 92 object categories, including people and common household items, across diverse environments such as offices, streets, and supermarkets, etc. The dataset intentionally captures severe visual ambiguity, with an average object occlusion rate of 26.6\%.

\subsection{Evaluation Protocol and Metrics}
\label{ssec:eval_protocol}

We evaluate our model's counting fidelity using a held-out test set of densely populated prompts. The generated images are subjected to an automated strict evaluation protocol leveraging zero-shot detection pipelines (SAM 3 \cite{carion2025sam3segmentconcepts}) tuned for high precision. To ensure the reliability of this automated evaluation and mitigate potential biases, we validate the SAM 3 detections against manual human annotations on a subset of the data. Our primary metrics include:
 \textbf{Exact Match Accuracy (\%):} The percentage of object categories where the predicted count exactly matches the target count.
\textbf{Relative Tolerance Accuracy ($\pm 10\%, \pm 20\%$):} The percentage of object categories whose predicted count falls within a strict percentage-based error margin, scaling naturally with large object counts.
    \textbf{Error Magnitude (MAE and RMSE):} Mean Absolute Error measures the average absolute deviation between target and detected counts per category, while Root Mean Square Error disproportionately penalizes massive outlier failures (such as the fusion of many objects).

\FloatBarrier
\subsection{Comparison Against State-of-the-Arts}


\begin{table}[htbp]
  \centering
  \caption{\textbf{OverlapDepth-45K} ($1{,}000$-image test): T2I vs.\ L2I under a fair layout protocol. Layout rows use identical \textbf{LLM-inferred} boxes and labels generated by Gemini 3 Flash (\cref{sssec:t2i_l2i_baseline_compare}).
  }
  \label{tab:overlap_t2i_l2i}
  \resizebox{0.8\linewidth}{!}{
  \begin{tabular}{@{} >{\raggedright\arraybackslash}p{4.2cm} >{\centering\arraybackslash}p{1.1cm} *{5}{>{\centering\arraybackslash}p{1.05cm}} @{}}
    \toprule
    & & \multicolumn{3}{c}{\textbf{Accuracy (\%) $\uparrow$}} & \multicolumn{2}{c}{\textbf{Error $\downarrow$}} \\
    \cmidrule(lr){3-5} \cmidrule(l){6-7}
    \textbf{Method} & \textbf{Type} & Exact & $\pm 10\%$ & $\pm 20\%$ & MAE & RMSE \\
    \midrule
    FLUX.2-Klein-4b (text only) & T2I & 13.24 & 18.18 & 30.59 & 5.21 & 9.96 \\
    SDXL-Turbo~\cite{podell2024sdxl} (text only) & T2I & 5.30 & 9.50 & 18.50 & 10.05 & 14.71 \\
    \midrule
    GLIGEN~\cite{gligen} & L2I & 8.84 & 15.48 & 29.15 & 11.41 & 18.76 \\
    MIGC~\cite{zhou2024migc} & L2I & 10.46 & 19.12 & 30.17 & 4.83 & 6.96 \\

    3DIS~\cite{3dis2025} & L2I & 12.88 & 19.48 & 30.17 & 4.07 & 6.34 \\
    \rowcolor{lightgray}
    \textbf{AIBL (Ours)} & \textbf{L2I} & \textbf{15.32} & \textbf{20.12} & \textbf{33.11} & \textbf{3.89} & \textbf{5.99} \\
    \bottomrule
  \end{tabular}
  }
  \vspace{-2mm}
\end{table}

\parag{Text-to-image vs.\ layout-to-image baselines on OverlapDepth-45K.}
\label{sssec:t2i_l2i_baseline_compare}
Benchmarking mixes two confounds: whether a model receives explicit spatial anchors, and \emph{which} boxes are supplied at test time. To ensure a fair comparison and separate paradigm differences from layout-quality bias, \cref{tab:overlap_t2i_l2i} contrasts \textbf{text-to-image (T2I)} systems that only observe the revised count prompts against \textbf{layout-to-image (L2I)} systems that additionally receive the same \emph{LLM-inferred layout} boxes (generated via Gemini 3 Flash) and category labels. This matches spatial conditioning across GLIGEN~\cite{gligen}, MIGC~\cite{zhou2024migc}, and 3DIS~\cite{3dis2025}, and uses the identical setup for our model. Automated counting follows \cref{ssec:eval_protocol}. 
As shown in \cref{tab:overlap_t2i_l2i}, our method consistently outperforms all L2I baselines across all metrics, yielding an absolute Exact Match improvement of +2.44\% over the strongest baseline (3DIS) and reducing RMSE from 6.34 to 5.99.

\begin{table}[htbp]
  \centering
  \caption{State-of-the-art comparison on CoCoCount and the T2I-Compbench counting subset. Best results are highlighted in \textbf{bold}.}
  \label{tab:sota_comparison}
  \resizebox{0.8\textwidth}{!}{
  \begin{tabular}{@{}lccc@{}}
    \toprule
    & \multicolumn{2}{c}{\textbf{CoCoCount}~\cite{makeitcount}} & \textbf{T2I-Compbench}~\cite{t2icompbench} \\
    \cmidrule(lr){2-3} \cmidrule(l){4-4}
    \textbf{Method} & \textbf{YOLOv9 Acc.} & \textbf{Human Acc.} & \textbf{Human Acc.} \\
    \midrule
    SDXL~\cite{podell2024sdxl} & 28 & 26 & 29 \\
    Repeated Object~\cite{hertz2022prompt} & 17 & 18 & 14 \\
    Reason Out Your Layout~\cite{lu2024reason} & 21 & 26 & 15 \\
    DALL-E 3~\cite{dalle3} & 25 & 38 & 36 \\
    Random masks + BoundedAttn~\cite{phung2023grounded} & 29 & 30 & 35 \\
    Counting Guidance~\cite{kang2025counting} & 21 & 22 & 22 \\
    CountGen (Make-It-Count)~\cite{makeitcount} & 50 & 54 & 48 \\
    \rowcolor{lightgray}
    \textbf{AIBL (Ours)} & \textbf{58} & \textbf{60} & \textbf{65} \\
    \bottomrule
  \end{tabular}
  }
  \vspace{-3mm}
\end{table}
\parag{Comparison on external benchmarks.}
To further validate the effectiveness of our approach, we evaluate our method on the counting subsets of T2I-Compbench~\cite{t2icompbench} and CoCoCount~\cite{makeitcount} against state-of-the-art methods, including recent specialized counting pipelines like CountGen~\cite{makeitcount}.
As shown in \cref{tab:sota_comparison}, we report baseline numbers as provided by Binyamin et al.\ \cite{makeitcount}. Our trained model significantly outperforms existing approaches. We randomly sample a subset of 100 images for human evaluation on T2I-Compbench. Under this evaluation, our approach achieves 65\% accuracy, establishing a massive absolute improvement of +17\% over CountGen (48\%) and +29\% over DALL-E 3 (36\%). This robust performance on complex, open-world prompts demonstrates that the principles of amodal gradient weighting and instance isolation generalize beyond our dense synthetic distributions.


\begin{figure*}[htbp]
  \centering
  \begin{minipage}{0.8\textwidth}
    \centering
    \includegraphics[
      width=\linewidth,
      trim=0 2mm 0 1.5cm,
      clip
    ]{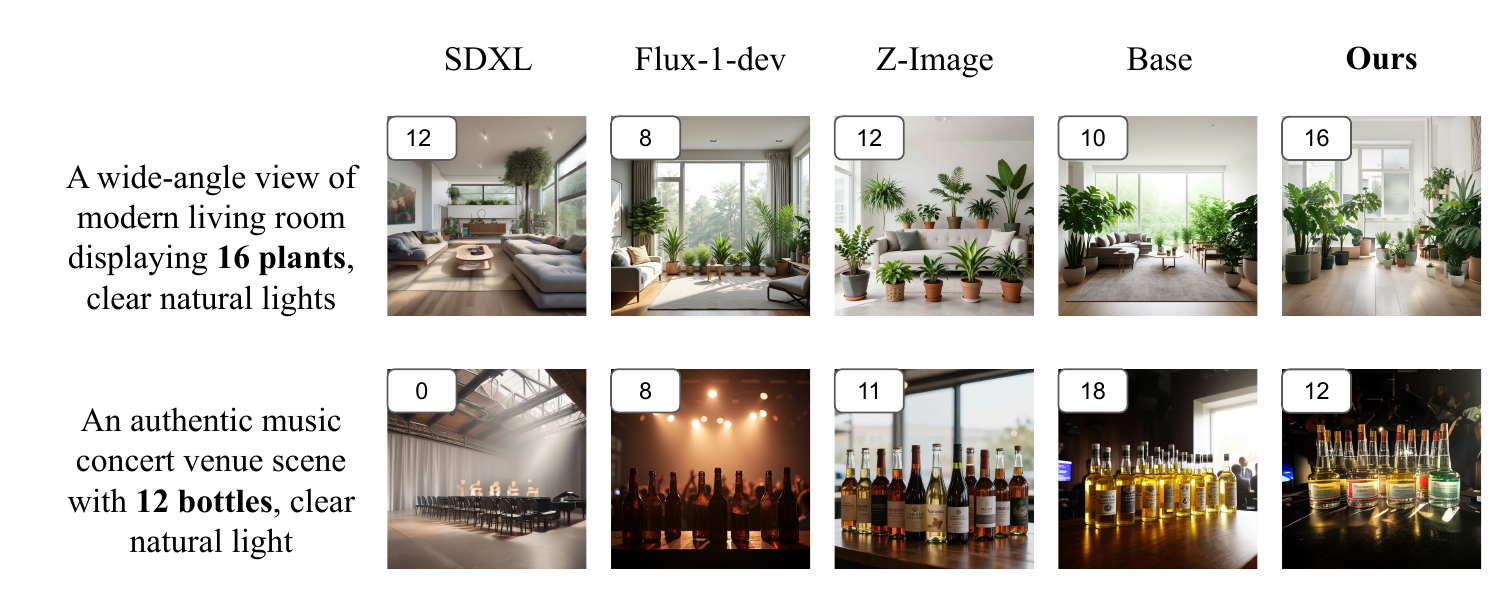}
    \vspace{-1mm}

    {\small
    \makebox[0.25\linewidth][c]{        }%
    \makebox[0.13\linewidth][c]{SDXL}%
    \makebox[0.165\linewidth][c]{Flux-1-dev}%
    \makebox[0.165\linewidth][c]{Z-Image}%
    \makebox[0.13\linewidth][c]{Base}%
    \makebox[0.165\linewidth][c]{Ours}%
    }
  \end{minipage}

  \caption{\textbf{Qualitative comparison on counting prompts.} Even with explicitly grounded bounding boxes, the layout-attention baseline (Base) frequently merges nearby instances. Our Amodal-Aware Instance-Balanced Loss (AIBL) rebalances training to promote clearer instance separation, producing more distinct objects (Ours). We also include reference generations from SDXL, FLUX.1-dev~\cite{blackforestlabs2024announcingbfl}, and Z-Image~\cite{team2025zimage} on the same prompts.}
  \label{fig:qualitative}
  \vspace{-3mm}
\end{figure*}

\parag{Qualitative Evaluation}
\label{ssec:qualitative}

We visually compare generations from the layout attention baseline against our Layout Attn.\ + AIBL model in \cref{fig:qualitative}. The baseline exhibits pronounced ``feature diffusion'' in dense regions - fusing canine bodies into a monolithic mass (\textit{11 dogs}) or merging pots into indistinct green blobs (\textit{16 plants}) -rather than generating discrete objects. In contrast, AIBL restores strict instance boundary separation: because heavily occluded regions receive proportionally stronger training gradients, the model preserves structural integrity even in constrained areas, resolving overlapping bottles, plants, and animals as countable entities.


\FloatBarrier
\subsection{Ablation Study}
\label{ssec:results}

We conduct ablation studies on the OverlapDepth-45K test set to isolate the contribution of each component. We evaluate the following configurations: (1) \textbf{Untrained (FLUX.2)}: the unconditioned baseline without layout training, (2) \textbf{Basic Layout Attn.}: layout attention without overlap-ownership bias, (3) \textbf{Basic Layout Attn. + AIBL ($\gamma{=}0.5$)}: adding occlusion-aware loss reweighting, and (4) \textbf{Ours (Full)}: our complete model adding the overlap-ownership bias. All trained models use identical LoRA rank, learning rate ($5 \times 10^{-5}$), and training duration (3000 steps).

\begin{table}[htbp]
  \centering
  \caption{Counting accuracy on the OverlapDepth-45K test set. Exact denotes Exact Match Accuracy while $\pm 10\%$ and $\pm 20\%$ denote Relative Tolerance Accuracy. Best results are highlighted in \textbf{bold}.}
  \label{tab:results}
  \resizebox{0.9\linewidth}{!}{
  \begin{tabularx}{\linewidth}{@{} >{\raggedright\arraybackslash}p{5.2cm} *{5}{>{\centering\arraybackslash}X} @{}}
    \toprule
    & \multicolumn{3}{c}{\textbf{Accuracy (\%) $\uparrow$}} & \multicolumn{2}{c}{\textbf{Error $\downarrow$}} \\
    \cmidrule(lr){2-4} \cmidrule(l){5-6}
    \textbf{Model} & Exact & $\pm 10\%$ & $\pm 20\%$ & MAE & RMSE \\
    \midrule
    Baseline & 13.24 & 18.18 & 30.59 & 5.21 & 9.96 \\
    Basic Layout Attn. & 14.71 & 18.24 & 31.18 & 4.47 & 9.26 \\
    \quad + AIBL ($\gamma{=}0.5$) & \textbf{15.34} & 19.21 & 32.12 & 4.22 & 8.12 \\
    \quad + Overlap-Ownership (Ours) & 15.32 & \textbf{20.12} & \textbf{33.11} & \textbf{3.89} & \textbf{5.99} \\
    \bottomrule
  \end{tabularx}}
\end{table}

\parag{Effectiveness of components.}
As shown in \cref{tab:results}, integrating basic layout attention substantially improves counting over the untrained baseline. Adding AIBL further amplifies these gains. Finally, applying our Overlap-Ownership bias ($\gamma{=}0.5$) achieves the strongest results on nearly every metric, with RMSE dropping to 5.99 - a massive 35.3\% reduction from the basic layout-only baseline (9.26). Because RMSE penalizes large-magnitude errors quadratically, this reduction provides compelling evidence that AIBL mitigates \textit{catastrophic failure modes}: the severe object merging in dense layouts that causes massive undercounting spikes. Crucially, this configuration improves nearly \emph{all} metrics simultaneously, demonstrating that occlusion-aware reweighting not only eliminates catastrophic failures but also provides a broadly beneficial training signal.

\begin{figure*}[htbp]
  \centering
  \begin{minipage}[t]{0.48\textwidth}
    \centering
    \includegraphics[width=0.7\linewidth]{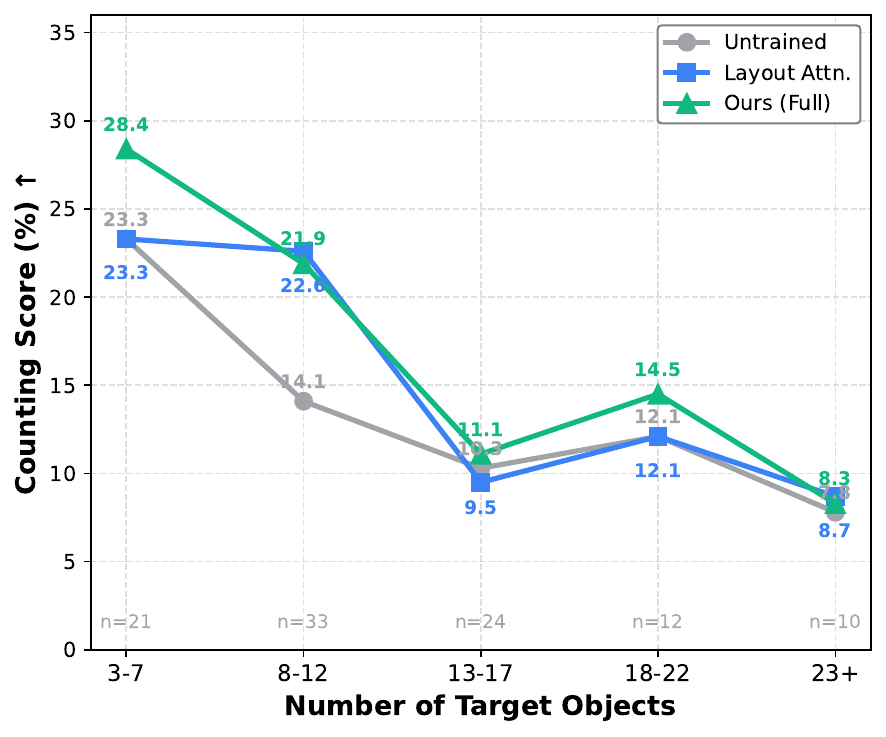}
  \end{minipage}\hfill
  \begin{minipage}[t]{0.48\textwidth}
    \centering
    \includegraphics[width=0.7\linewidth]{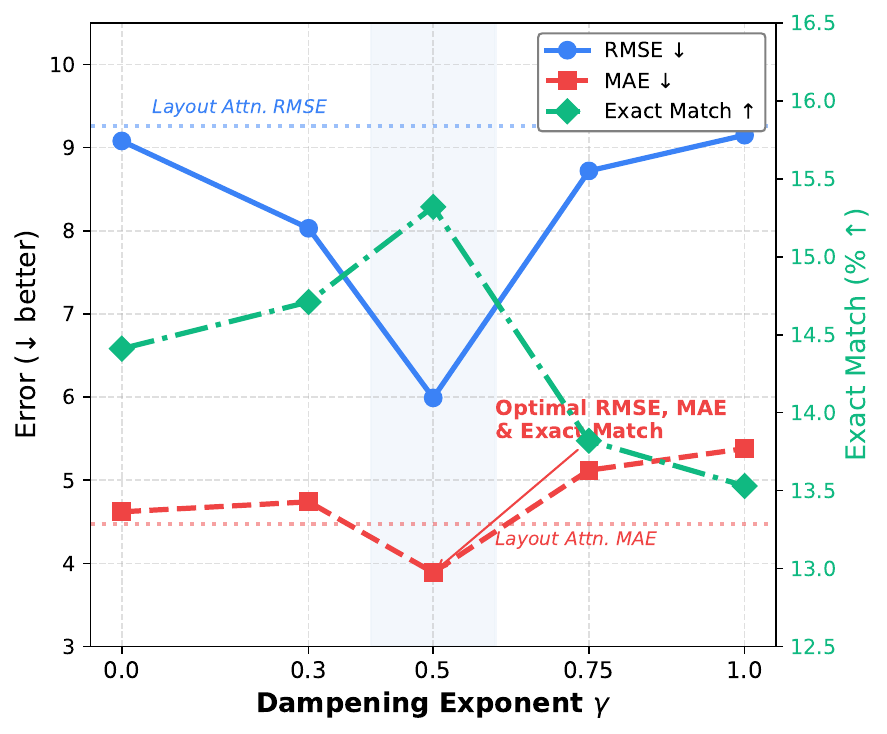}
  \end{minipage}
  \vspace{-2mm}
  \caption{\textbf{Analysis of AIBL behavior.} (\textit{Left}) Counting score stratified by target object density. AIBL provides significant improvements in mid-to-high density regimes (13--22 objects). (\textit{Right}) RMSE, MAE, and Exact Match accuracy as a function of the dampening exponent $\gamma$. Dotted lines indicate the layout-only baseline. The optimal $\gamma{=}0.5$ achieves the best trade-off across all metrics.}
  \label{fig:aibl_analysis}
\end{figure*}

\parag{Counting Score by Scene Density.}
We further evaluate the \textit{counting score} - the percentage of target objects correctly generated and detected - stratified by scene density (\cref{fig:aibl_analysis}a). Unlike binary Exact Match, this metric captures partial success (e.g., 18/20 cups = 90\%). AIBL provides its strongest improvements in the mid-to-high density regimes (13 - 22 objects), where occlusion is most severe: in the 18 - 22 bin it achieves 14.5\% vs.\ 12.1\% for layout-only (19.8\% relative gain), confirming that gradient rebalancing is most impactful where occluded instances are most likely to be dropped. Even in the low-density regime (3 - 7 objects), AIBL delivers a 5.1 percentage-point gain, suggesting that amodal-aware weighting helps preserve instance boundaries across densities. In the highest bin (23+ objects) the layout-only baseline marginally outperforms AIBL (8.7\% vs.\ 8.3\%), likely because extreme overlap exceeds the regime where gradient boosting alone can resolve all instances - an open challenge for future work.

\parag{Sensitivity to dampening exponent $\gamma$.}
\label{ssec:gamma_ablation}
The dampening exponent $\gamma$ in \cref{eq:alpha} controls the aggressiveness of occlusion-based reweighting: smaller values compress the boost range, while larger values preserve the raw amodal-to-visible ratio but risk gradient instability. We evaluate $\gamma \in \{0.0, 0.3, 0.5, 0.75, 1.0\}$ with all other hyperparameters fixed. Results reveal a clear inverted-U pattern (\cref{fig:aibl_analysis}b). At $\gamma{=}0$, AIBL reduces to foreground-vs-background reweighting, yielding only a modest RMSE improvement (9.08 vs.\ 9.26 for layout-only). The optimal $\gamma{=}0.5$ achieves the best results across nearly \emph{all} metrics simultaneously, striking the ideal balance between gradient amplification and training stability. Beyond this point, performance degrades: at $\gamma{=}1.0$, a 90\%-occluded object receives a $10\times$ raw boost, and the resulting RMSE (9.15) approaches the layout-only baseline, effectively negating the benefits of occlusion-aware reweighting.

\begin{figure*}[htbp]
  \centering
  \includegraphics[width=0.8\textwidth, trim=0 1.6cm 0 0, clip]{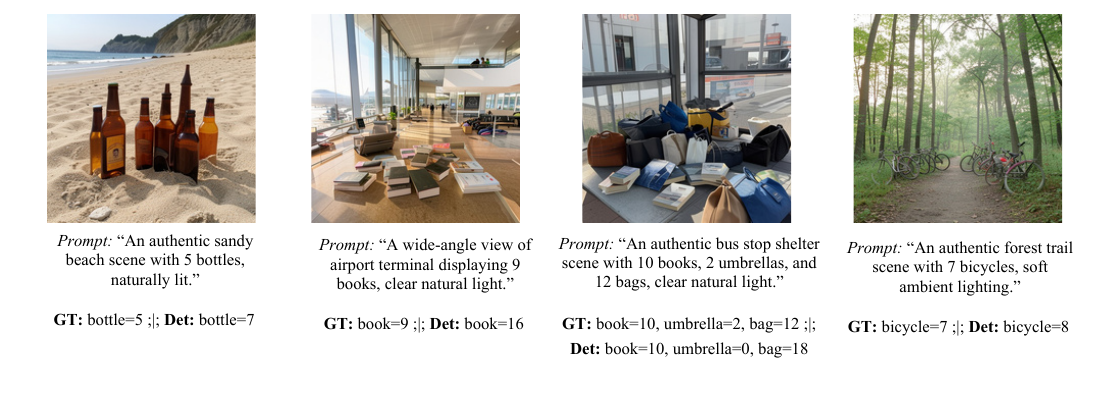}
  \vspace{-3mm}
  \caption{\textbf{Failure cases.} Our method can struggle with extreme density or thin objects. }
  \label{fig:failure_cases}
  \vspace{-5mm}
\end{figure*}

\subsection{Discussion and Limitations}

While AIBL effectively mitigates object fusion, three inherent limitations remain (\cref{fig:failure_cases}). (1) \emph{Density saturation}: In extremely packed scenes, latent space resolution limits force feature overlap, causing the model to merge nearby objects despite balanced gradients. (2) \emph{Category bias}: Thin or articulated objects (e.g., umbrellas) lose structural detail during latent compression, making them harder to render than simpler shapes (e.g., bags), inflating per-category errors. (3) \emph{Synthetic domain shift}: OverlapDepth-45K is entirely synthetic. While zero-shot evaluations on real datasets (e.g., T2I-CompBench) show strong generalization, synthetic scenes can encode non-canonical lighting and artifactual layouts. Future work should explore explicit shape priors, category-aware loss reweighting, and real-world hybrid datasets to overcome these challenges.

\section{Conclusion}

In this paper, we introduced Amodal-Aware Instance-Balanced Loss (AIBL), a novel training-time reweighting scheme that dynamically scales the loss contribution of each instance based on the ratio of its amodal extent to its visible area. Coupled with a distance-weighted grounded layout attention mechanism and our dedicated OverlapDepth-45K benchmark, AIBL amplifies supervisory gradients for heavily occluded objects without introducing inference-time mask dependencies or training-inference mismatch. Experimental results show that our proposed method effectively mitigates catastrophic object fusion, while strictly preserving baseline photorealism and image quality. Future work will focus on overcoming this extreme-density overlapping objects, potentially drawing inspiration from spatial priors that explicitly reason about 3D occlusion to better disentangle complex semantic overlaps.

\bibliographystyle{plain}
\bibliography{main}






\end{document}